# A Three-Stage Algorithm for the Large -Scale Dynamic Vehicle Routing Problem with Industry 4.0 Approach

Maryam Abdirad[1], Krishna Krishnan[1], Deepak Gupta[1]

## Abstract

Companies are eager to have a smart supply chain especially when they have a dynamic system. Industry 4.0 is a concept which concentrates on mobility and real-time integration. Thus, it can be considered as a necessary component that has to be implemented for a Dynamic Vehicle Routing Problem. The aim of this research is to solve large-scale DVRP (LSDVRP) in which the delivery vehicles must serve customer demands from a common depot to minimize transit cost while not exceeding the capacity constraint of each vehicle. In LSDVRP, it is difficult to get an exact solution and the computational time complexity grows exponentially. To find near optimal answers for this problem, a hierarchical approach consisting of three stages: "cluster-first, route-construction second, route-improvement third" is proposed. The major contribution of this paper is dealing with large-size real-world problems to decrease the computational time complexity. The results confirmed that the proposed methodology is applicable.

**Keywords:** Dynamic Vehicle Routing problem, Clustered Vehicle Routing problem, Three-stage algorithm, Industry 4.0

## Introduction

The vehicle routing problem (VRP) is one of the well-known supply chain problems. This problem was defined by Dantzig. The goal of this problem is to minimize the total transportation cost and seek a route to deliver demands to customers. There are different varieties of the VRP [1][2][3], for example, Capacitated Vehicle Routing Problem (CVRP) [4], VRP with time windows (VRPTW) [5], and Multi-Depot Vehicle Routing Problem (MDVRP) [6]. Recently, the dynamic vehicle routing problem (DVRP), one of the varieties of VRP, has got significant consideration.

Corresponding author:
Maryam Abdirad
mxabdirad@wichita.edu
Krishna Krishnan
Krishna.Krishnan@wichita.edu
Deepak Gupta
Deepak.gupta@wichita.edu
[1] Department of Industrial, Systems, and Manufacturing Engineering, Wichita State University, Wichita, KS 67260, USA

The DVRP has dynamic demands that arrive in the system at different times. These demands obviously affect the solution because they change both the problem and the solution the instant they arrive in the system. The challenge of this subject is to find a route from a depot to the destination with respect to minimum distances.

Most real-world DVRPs are large and complex. Most companies are enthusiastic is to establish a modern supply chain (MSC) that is effective, automated, transformative, and comprehensible, which can solve a DVRP with a high volume of data. Industry 4.0 prepares a framework that can manage moves from a conventional supply chain to an MSC by focusing on increasing automation, digitalization, interconnection in companies through Internet of Things [IoT] and cyber-physical systems. The functionality of the DVRP perfectly matches the concepts of Industry 4.0. Therefore, Industry 4.0 can be used as an infrastructure for a DVRP [7]–[12]

In an Industry 4.0 environment, to analyze the generated data, different machine learning techniques and artificial intelligence (AI) algorithms may be applied. AI can help to solve problems faster than an exact solver by reducing the computational time as well as problem complexity [13]. One example is the usage of AI algorithms to optimize the supply chain and manufacturing operations, to help manufacturing operations respond better and faster to anticipated changes in the market. In this research, it was decided to use these techniques in solving the DVRP.

The purpose of the research is to initiate a single depot dynamic vehicle routing problem in a large-scale demand network which is called the large-scale DVRP (LSDVRP). The VRPs are not solved in polynomial time, which in turn makes this problem into an NP-hard problem. In this work, a three-stage algorithm is proposed to solve the LSDVRP problem. In this approach, three clustering algorithms are used in the first stage. In the second stage, three different construction algorithms are used to develop the solutions for the different clusters. In the third stage, the solutions obtained from the construction algorithms are improved using three different improvement algorithms to identify the best solutions to the DVRP. The clustering method allows the large LSDVRP problems to be subdivided into smaller problems and this reduces the computational complexity and helps to solve very large problems. During the clustering phase of the solution procedure, customers are assigned to vehicles. The three different clustering algorithms used include: K-mean clustering, BIRCH (balanced iterative reducing and clustering using hierarchies) clustering, and Gaussian Mixture Models (GMM) algorithms. In the second and third stages, the VRP in each cluster is solved by a combination of heuristic algorithms consisting of construction algorithms (second phase) and improvement algorithms (third phase). Nine different cases are used to demonstrate the proposed solution approach. One of the most important contributions of this article is that the proposed hierarchical approach can deal with large size problems. In this work, for the first time, a combination of the clustering algorithms and the construction and improvement algorithms is introduced for solving the LSDVRP.

In the following sections, a brief literature review about the role of Industry 4.0 in MSC and the VRP, with prominence on DVRPs are provided. In Section 3, the problem statement is proposed. A solution approach is presented in Section 4. Different scenarios with experimental computations and results are presented in Section 5, and finally, conclusions and prospects for future work in section 6 conclude this paper.

## Literature review

### *Industry 4.0*

Originated in 2011 by Henning Kagermann [14], Industry 4.0 aims at developing a smart manufacturing network where materials, goods and machines, and products interconnect automatically with no human involvement [15]–[22]. This concept has emerged in the literature under different labels such as "smart manufacturing," "Fourth Industrial Revolution," "integrated industry" or "industrial internet" [15][23]. There is no deterministic definition for Industry 4.0. As said by Lopes de Sousa Jabbour et al. "the core feature of Industry 4.0 is connectivity between machines, orders, employees, suppliers, and customers due to the Internet of Things (IoT), and electronic devices; as a consequence, firms are able to produce products using decentralized decisions and autonomous systems"[24].

Industry 4.0 is realized when manufacturing systems and factories consist of smart sub-entities or tasks like smart machines, connected devices, sensors, actuators, processes, logistics, suppliers, and products [12], [25]–[28]. Therefore, Industry 4.0 advocates the adoption and application of the latest technologies to supply chain, manufacturing, and management for real-time connectivity. These technologies (digital transformation) include but are not limited to CPSs, cloud manufacturing, IoT, the Internet of Services (IoS), robotics, big data [29]–[32]. Integration of these technologies can substantiate the level of automation that Industry 4.0 aspires to achieve when used in tandem with computational techniques like artificial intelligence (AI), business analytics (BA), and dynamic optimization (DM).

By implementing Industry 4.0 in the supply chain systems also known as Supply Chain 4.0, four main SC elements— integration, operations, purchasing, and distribution—are affected and can increase the productivity of companies as well [33]. The literature suggests that Industry 4.0 can significantly improve supply chains, business models, and processes to achieve an MSC. Improvements are expected in the lead time for product shipment and delivery, in the efficiency of response to unforeseen events, and in decision-making quality [34]. Given the depth and breadth of its impact on SCs, Industry 4.0 enables companies to implement complicated and dynamic processes to manage large-scale production and integration of customers as well as operations planning and logistics [35][36]. With Supply Chain 4.0, information exchange occurs in real-time to drive decisions at different organizational levels. Accordingly, the presence of Industry 4.0 can promote efficiency in dynamic vehicle routing as a critical component of supply chain design and management.

### *Dynamic Vehicle Routing Problem and Clustering VRP*

The emergence of advanced communication and information technologies has engendered new opportunities for dynamic vehicle routing problem (DVRP) [37]. Originated in 1977 by Wilson and Colvin [38], the concept of DVRP gradually advanced to address the real-time of change of routes based on the receipt of immediate new requests by customers [39]. In fact, DVRP accommodates dynamic input data [40].

With the uptake of Industry 4.0, there is an opportunity to use connected sensors, devices, and positioning systems to coordinate SCs and its actants like control towers, depots, and drivers of the vehicles (Fig. 1). Therefore, this can enable actants to drive decisions like changing their delivery plans and routines based on the real-time location of drivers (with data from with global positioning system; GPS) and new demands from customers [41].

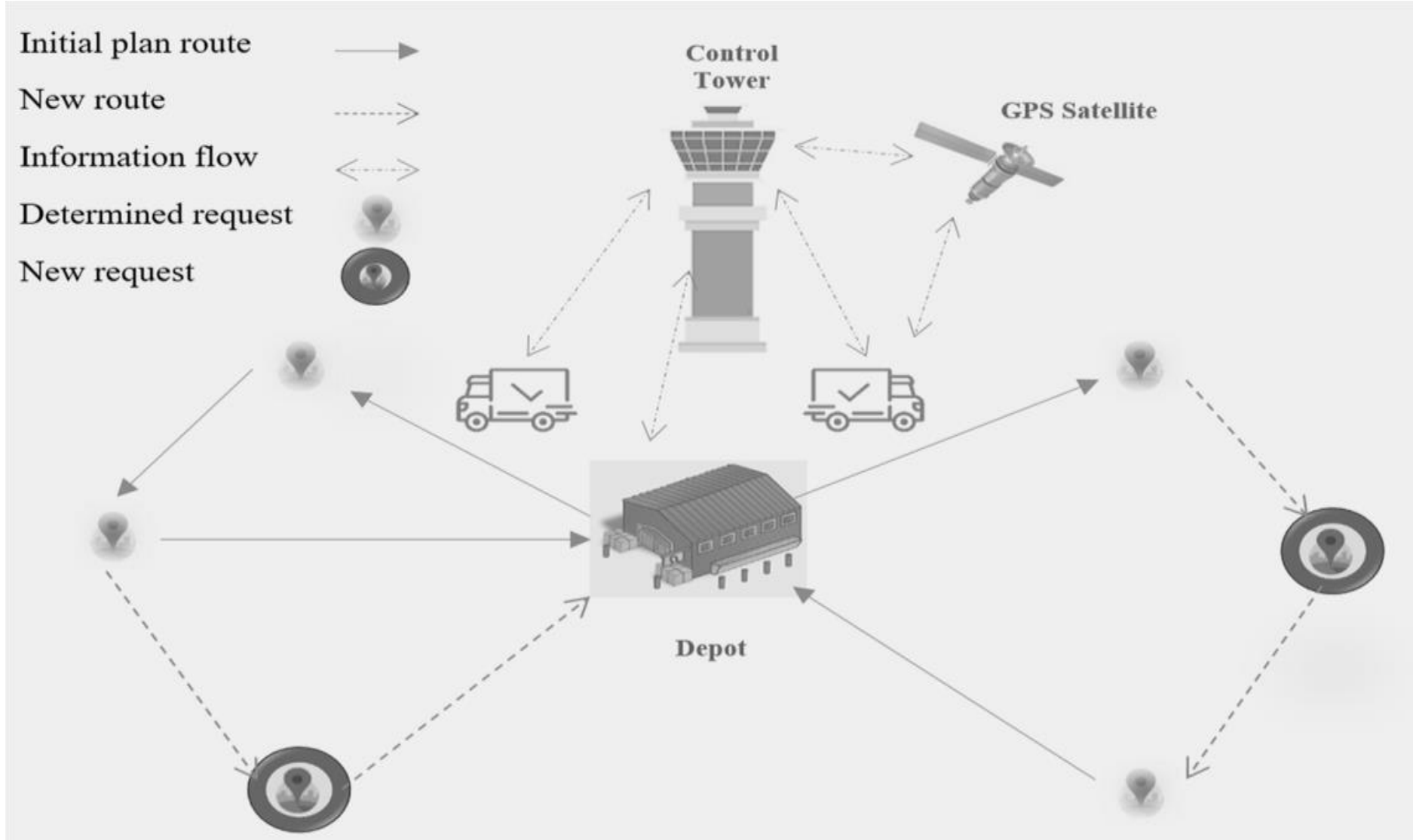

**Fig. 1.** Industry 4.0 facilitating communication for DVRP [42][43].

Recent advances in heuristics provide unprecedented technical advantages for tackling large scale VRP problems. To two types of heuristics include: (1) cluster first & route second, in which clusters of customers assigned to different vehicles and then routes are planned for each cluster; and (2) route first & cluster second, in which the solution builds a traveling salesman tour through all customers and then partitions the tour for vehicle assignment [44]. This research builds an innovative three-stage algorithm for DVRP: a cluster-first and route-construction second, followed by a route-improvement hierarchical method is applied.

There is extensive research related to VRP based on different constraints and types. In this section, only some of the studies related to Clustering VRP (CluVRP) and DVRP in the literature are summarized. Sevaux et al. introduced the clustering technique to divide the customer zones to deliver parcels. The result was then solved as a classical VRP using the zones [45]. However, it is hard to find an exact solution for this problem, a new exact algorithms for the CluVRP was introduced by Battarra, Erdoğan and Vigo [46].

The application of machine learning algorithms in solving the VRP appeared in some research. Korayem, Khorsid, and Kassem combined the Grey Wolf Algorithm (GWO) with the K-means clustering algorithm to generate the 'K-GWO' algorithm for the VRP [47]. Pop et al. focused on the clustered vehicle routing problem (CluVRP) which divided this problem into two logical and natural smaller subproblems: They established the first subproblem is to determine the (global) routes visiting the clusters using a genetic algorithm. The second subproblem is solved by transforming each global route into a traveling salesman problem (TSP) [48].

Nallusamy et al. used k- Means clustering, the given cities depending upon the number of vehicles and each cluster is allotted to a vehicle to solve multiple VRP (mVRP). After clustering and after the cities were allocated to the various vehicles, each cluster was taken as an individual Vehicle Routing problem and genetic algorithm (GA) was applied to each cluster. They found this approach gave a better result and a more optimal tour for mVRPs in short computational time than other Algorithms [49].

Dondo and Cerda solved single depot and multi-depot large scale VRPs by using the initial Clustering solution and then used the MILP problem formulation for each cluster [50]. Bujel et al. proposed to cluster nodes by using Recursive-DBSCAN clustering; their approach leads to a 61% decrease in runtimes of the CVRPTW solver against Google Optimization Tools [51]. Xu, Pu, and Duan, solved this problem by K-means that divided the region based on distance; then an enhanced Ant Colony Optimization (E-ACO) heuristic handles each region [52]. Gharib et al. proposed a two-stage algorithm for times of crisis. First, the local distribution region of different distribution vehicles is obtained by the Fuzzy C-Means Clustering Algorithm. Then, they optimize the path of different distribution regions based on the Ant Colony Algorithm [53].

Barthelemy et al. proposed a two-stage algorithm designed by the Clarke and Wright heuristic along with a 2-opt local search in the first step. The result from the first step is used as input for the second stage, which is simulated annealing (SA) along with a 2-opt local search procedure [54]. Özdamar and Demir developed a hierarchical cluster and route procedure (HOGCR) for coordinating vehicle routing in large-scale post-disaster distribution and evacuation activities. This method is a multi-level clustering algorithm that groups demand nodes into smaller clusters at each planning level until the final cluster sizes enable the optimal solution of the cluster networks' routing problems, thus enabling the optimal solution of cluster routing problems [55]. Comert et al. proposed a two-stages algorithm, "cluster-first route-second". In the first stage, customers are assigned to vehicles using three different clustering algorithms of K-means, K-medoids, and DBSCAN. In the second stage, a VRP with time window problem is solved using a MILP exact method [56]. In a follow-up paper [57], for the first stage, they used K-means, K-medoids, and random clustering algorithms. In the second stage, routing problems for each cluster were solved using a branch and bound algorithm. Both methodologies were employed on a case study in a supermarket chain.

We have done extensive literature review and there is lack of research that solve the DVRP in real time. Hence, the focus of this research is to investigate the development of algorithms that can solve LSDVRP problems. In the next section, the problem definition is detailed.

## Problem Definition

LSDVRP consists of 'n' number of customers with known demand $d_i$ ($i = 1, \ldots, n$ – where n is the number of customers) and 'm' number of vehicles with fixed equal capacity, $q_i$ ($i = 1, \ldots, m$). Vehicles start from a depot to deliver products to the customers and end at the depot. Meanwhile, new customers with known demand submit their product demands dynamically over time. There is an assumption that the sum of the demands in each vehicle route does not exceed the vehicle's capacity. The distance of the route is given by the Euclidian distance, between customers' locations. Other given constraints are as follows:

- All customer demands should be completely accepted.
- Customer demands should be fully fulfilled.
- Only one vehicle is assigned to each route.
- The cost of travel is directly proportional to the distance.

## Methodology

Even though, it is possible to solve small size VRPs, by an exact algorithm in a reasonable time period, LSDVRP should be solved by a metaheuristic approach. The present proposed methodology when the problem is large, is a cluster-first, construction-route second, improvement-route third method. In the clustering phase, customers are grouped into feasible clusters. A feasible cluster requires that the total customers' demands within the cluster do not exceed the capacity of a single-vehicle. In the second stage, the set of customers to be satisfied and the sequence in which they will be served by each vehicle within each cluster are identified, which allows this phase to be solved as a traveling salesman problem TSP [12]. This second stage is followed by an improvement phase in which the solutions are further enhanced. The steps are detailed below.

*Clustering:* The first stage of this methodology is clustering. Clustering analysis is an unsupervised learning heuristic that separates data into clusters or groups. One of the main aspects of clustering is to find customers with similar orders and in similar geographical locations. Based on the distribution of the customer data points, a suitable clustering algorithm should be selected. At first, the number of clusters is determined based on the number of available vehicles. After determining the number of clusters, the clusters are created using the selected clustering algorithm. The total number of demands of each cluster is calculated and controlled such that the capacity constraint is satisfied. The clusters that satisfy a capacity constraint are accepted as favorable clusters. There are different clustering algorithms that are applicable at this stage: K-means, Gaussian Mixture Models (GMM), Mean Shift Clustering, Hierarchical Clustering, etc. In this study, K-means, GMM, and BIRCH clustering algorithms are applied as clustering methods for solving this problem. If the clustering algorithms deliver a feasible result, then construction algorithms will be executed.

*Construction Algorithms:* The second-stage of the methodology is to construct a feasible route for each cluster from the first-level with an objective to minimize the total distance traveled. Heuristic algorithms are used to get a solution in a reasonable time. This stage includes Path Cheapest Arc (PCA), Global Cheapest Arc (GCA) and Savings algorithms. In this research, the best answer corresponds to the lowest value obtained for the distance traveled.

*Improvement Algorithms:* In this stage, the solutions obtained in the second stage are further improved. It is possible that the solutions obtained in the second phase may not be the best solutions and can be further improved. Therefore, to enhance the solution, improvement algorithms are applied. Improvement algorithms used in this phase of the methodology include Guided Local Search (GLS), Tabu Search (TS) and Simulated Annealing (SA). "Inter-route improvement" including exchanging or inserting between and during the routes can improve the routes [40].

After executing the three-stage algorithm, vehicles start delivering products to the customers. It is necessary to modify the vehicle routes when new customers enter the system (with new orders). Thus, this algorithm is re-executed each time a new order is submitted. However, all customers that have been served at that instant are removed from the system and the vehicle positions are also updated. This process of rerouting will occur every time a customer enters the system until all customers are served. Fig. 2 illustrates a visual representation of this methodology.

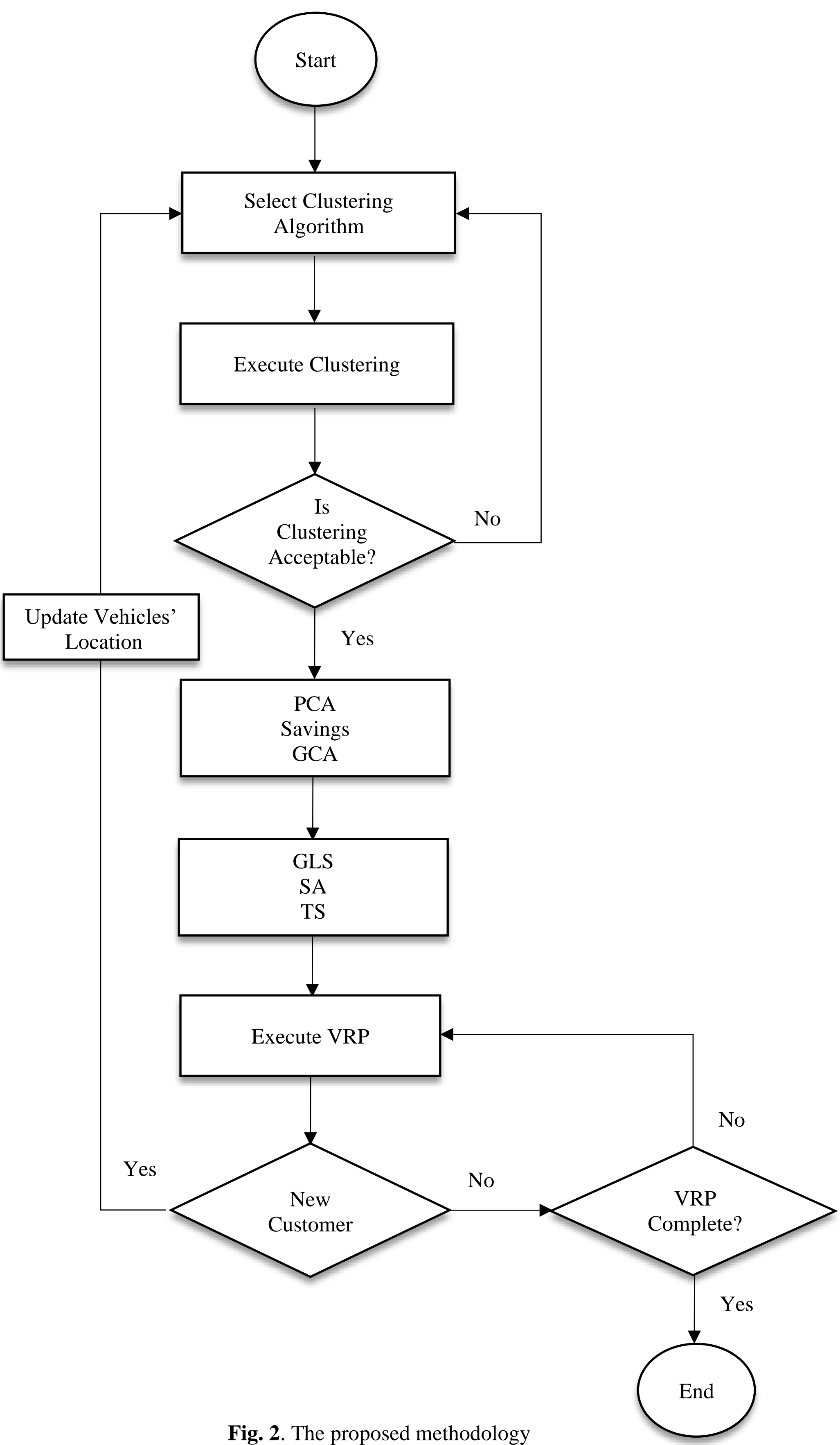

**Fig. 2**. The proposed methodology

The problem and solution approach are explained using a small example. In this example, there are four vehicles. At time t = 0, there are 11 customers in the system (Fig. 3). For the 4-Vehicle, 11-customer initial problem, clustering methods are used to assign customers to the vehicle. The number of clusters will always be equal to the number of vehicles (K) available in the system. Three different clustering approaches were used in stage 1 to identify the best clustering technique. The capacity of the vehicle is constrained to be greater than the total demand in each cluster. In the second stage, three different heuristic algorithms were applied to the clusters to determine a feasible initial route. The heuristics applied in the second stage are all construction heuristics. The solution obtained from the construction heuristics is further refined using improvement heuristics in stage 3.

At time t = 5, a new customer arrives into the system. However, the vehicles may have left the depot and is no longer at the home position. Based on the updated vehicle positions, and after eliminating the customers that were already served, the clustering (stage 1) is performed again. The construction of the new routes (stage 2) based on the cluster and the improvement of the route (stage 3) is also completed. This procedure is repeated any time a new customer enters the system, thus making the vehicle routing to be dynamic.

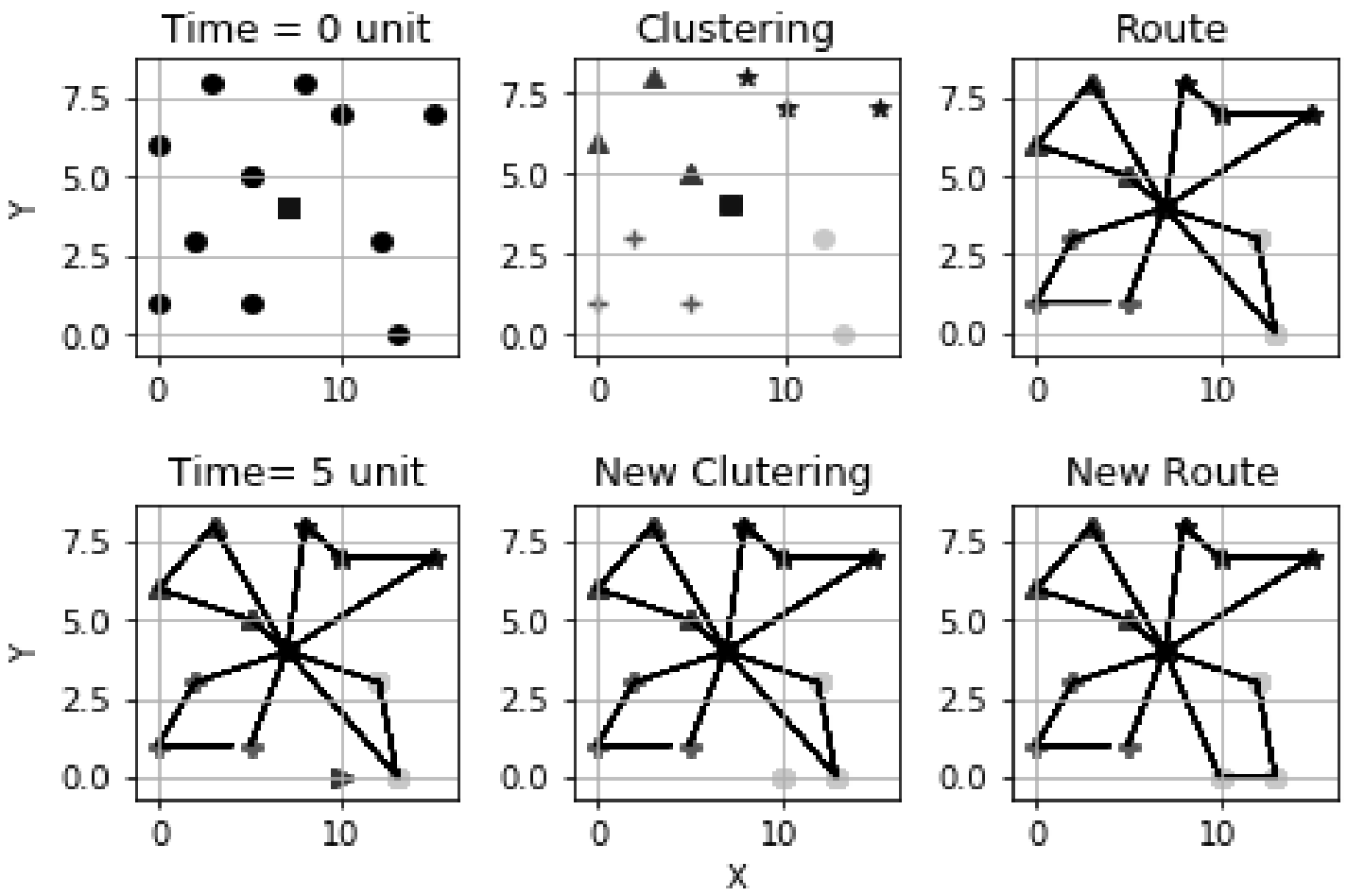

**Fig. 3.** The Three-Stage Algorithm proposed method.

## Experimentation and Discussion

To verify the efficiency of the proposed method, the proposed algorithm was implemented, and validate through several experiments. The experimental results are shown and discussed.

## *Data Collection and Processing*

There is no benchmark test problem available for a LSDVRP. The proposed method is tested with three different sizes of data sets: three small sample size data set and three large sample data set. The small sample size case studies include a single-depot, four-vehicle problem with a capacity of 70, and a total of 100 customers at time t = 0. The number of new customers (dynamic) during the time period of study is 20, 50, and 70 (3 different cases). These new customers enter the system randomly. The medium sample size case is also a single-depot, four-vehicle problem with a vehicle capacity of 125 units. At time t = 0, there are 200 customers in the system. However, the number of the dynamic or new customers during the study period is 120, 150, 170 (3 different cases). The large sample size case studies include a single-depot, four-vehicle problem with a capacity of 70, and a total of 300 customers at time t = 0. The number of new customers (dynamic) during the time period of study is 220, 250, and 270 (3 different cases).

The algorithms were implemented using Python. To validate the solution obtained, the three-stage algorithms is executed 10 times. The number of repetitions is based on the calculation of the deviation in the solution at 95% confidence interval. In most of the case results, the averages and minimum are equal. For the rest, the highest standard deviation among all case results is 5.3.

To verify the proposed method, three different clustering methods, K-means, Gaussian Mixture Models (GMM), and BIRCH clustering algorithms, which are based on the distance between points, were selected for stage 1. The reasons for selecting these algorithms are that these clustering techniques are well suited for the generated data for this problem and delivered good results. In addition, these clustering algorithms allow the user to specify the number of clusters, which for this research must be equal to the number of vehicles. Therefore, it is possible to limit cluster sizes. Previous researchers have used K-means and Gaussian methods for the vehicle routing problem and have been shown to generate good solutions for VRP [56]. The BIRCH algorithm however is new and is being tested along with the other two for its applicability to DVRP. Fig. 4. shows the results of the clustering algorithms for 100 points.

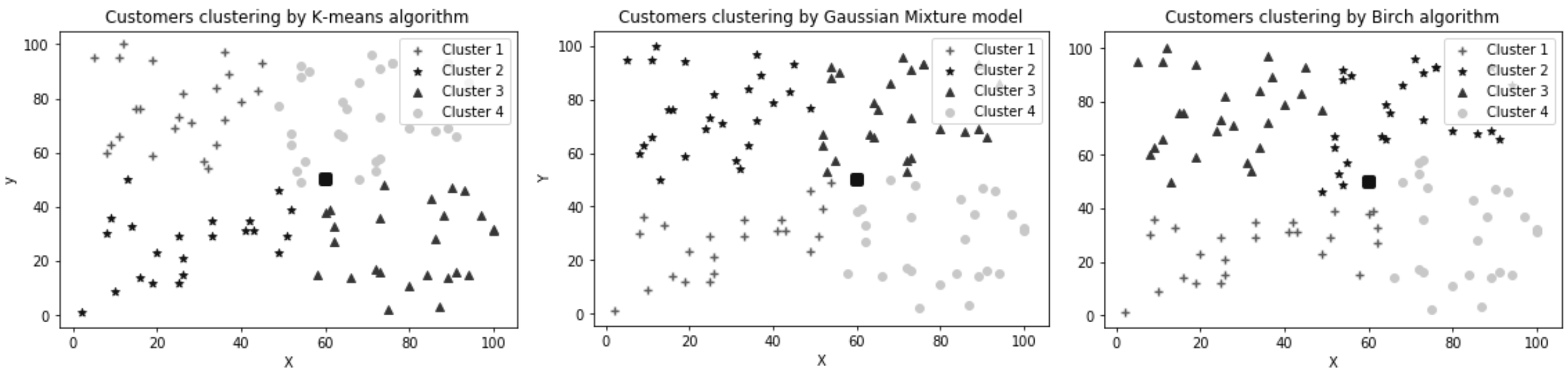

**Fig. 4.** Comparison of results for different clustering methods.

### *Experimental Result*

At first, the construction algorithms are applied to the initial data without performing clustering. Then the construction algorithms (Stage 2) were applied to the clusters obtained in Stage 1. The results of the stage_1-stage_2 algorithms are compared with the solution obtained by directly applying the construction algorithm to the data. Table 1 shows the costs obtained by using various algorithms. It also calculates the percentage improvement for the stage_1-stage_2 algorithm compared to the direct use of the Stage 2 algorithms on the raw data.

As shown in Table 1, for 100 initial customers and 20 dynamic customers, the K-means clustering and Savings algorithm resulted in a minimum cost of $949.6. When the number of dynamic customers is increased to 50, the K-means clustering and Savings algorithm resulted in a minimum cost of 1063.7. For the 100 initial customers with 70 dynamic customers, the K-means clustering and GCA algorithm resulted in the minimum cost of 1113.1.

For 200 initial customers and 120 dynamic customers, K-means clustering and GCA resulted in a minimum cost of 1853.8. By having 100 initial customers and increasing dynamic customers to 150, K-means clustering and GCA resulted in a minimum cost of 1963.1. For 200 initial customers and 170 dynamic customers , K-means clustering and PCA resulted in a minimum cost of 2132.

For 300 initial customers and 220 dynamic customers, K-means clustering and Savings algorithm resulted in a minimum cost of 2315.3. By having 300 initial customers and increasing dynamic customers to 250, K-means clustering and PCA resulted in a minimum cost of 2455.9. For 300 initial customers and 270 dynamic customers , K-means clustering and GCA resulted in a minimum cost of 2587.1.

It can be seen from Table 1 that applying the construction algorithms without clustering results in higher costs than the best-combined clustering and route construction algorithms. Also, between a combination of clustering algorithms and an initial algorithm, K-means clustering shows the best results.

**Table 1.** Computational results of the initial algorithms with clustering algorithm and initial algorithms.

| Dataset | Initial Customers | Dynamic Customers | Initial Algorithm | No Clustering | Clustering Algorithm | Clustering Algorithm and Initial Algorithm | |
|---|---|---|---|---|---|---|---|
| | | | | Average Cost ($) | | Average Cost ($) | Improvement (%) |
| Small size | 100 | 20 | Savings | 1142.7 | K-means | 949.6 | 16.90 |
| | 100 | 50 | Savings | 1239.1 | K-means | 1063.7 | 14.16 |
| | 100 | 70 | GCA | 1277.8 | K-means | 1113.1 | 12.89 |
| Medium size | 200 | 120 | GCA | 2032.8 | K-means | 1853.8 | 8.81 |
| | 200 | 150 | GCA | 2101.5 | K-means | 1963.1 | 6.59 |
| | 200 | 170 | PCA | 2218.5 | K-means | 2132 | 3.90 |
| Large Size | 300 | 220 | Savings | 2543.9 | K-means | 2315.3 | 8.99 |
| | 300 | 250 | PCA | 2601.3 | K-means | 2455.9 | 5.59 |
| | 300 | 270 | GCA | 2739.7 | K-means | 2587.1 | 5.57 |

Table 2 shows the result of performing the three-stage methodology which includes clustering using K-Means, BIRCH and GMM clustering techniques followed by a construction algorithm and an improvement algorithm. The results of each clustering technique along with corresponding construction and improvement algorithms are compared with the Two-Stage Algorithm.

**Table 2**. Computational results of three-stage algorithms and percentages improvement.

| **Dataset** | **Initial Customers** | **Dynamic Customers** | **Two-Stage Algorithm** | **No Clustering** | **Three-Stage Algorithm** | **Three-Stage Algorithm** | |
|---|---|---|---|---|---|---|---|
| | | | | **Average Cost ($)** | | **Average Cost ($)** | **Improvement (%)** |
| Small size | 100 | 20 | GCA and GLS | 990.2 | K-means, PCA and GLS | 930.6 | 6.02 |
| | 100 | 50 | PCA and GLS | 1178.7 | K-means, PCA and GLS | 1073.3 | 8.94 |
| | 100 | 70 | PCA and GLS | 1155.6 | K-means, PCA and GLS | 1029.1 | 10.95 |
| Medium size | 200 | 120 | GLS and TS | 1903.3 | K-means, GCA and TS | 1769.8 | 7.01 |
| | 200 | 150 | Savings and GLS | 2078.4 | K-means, Savings and GLS | 1906.9 | 8.25 |
| | 200 | 170 | PCA and GLS | 2173.5 | K-means, PCA and GLS | 2092 | 3.75 |
| Large size | 300 | 220 | PCA and GLS | 2481.3 | K-means, GCA and TS | 2275.9 | 8.28 |
| | 300 | 250 | PCA and GLS | 2520.5 | K-means, PCA and GLS | 2410.4 | 4.37 |
| | 300 | 270 | GCA and GLS | 2680.7 | K-means, PCA and GLS | 2539.3 | 5.27 |

By look at the small size problems, the lowest cost 100 initial customers and 20 dynamic customers when using the K-Means clustering technique is obtained when combined with the GCA and GLS algorithm at a cost of 930.6 and 6.02% improvement compared to Two-Stage Algorithm. The modified routes obtained when new customers enter the systems is shown in - Appendix Fig. 5. The lowest cost 100 initial customers and 50 dynamic customers when using the K-Means clustering with the PCA and GLS algorithm at a cost of 1073.3 and 8.94% improvement. The lowest cost 100 initial customers and 70 dynamic customers when using the K-Means clustering technique is with the PCA and GLS algorithm at a cost of 1029.1 and 10.95% improvement.

For medium size problems, the lowest cost 200 initial customers and 120 dynamic customers when using the K-Means clustering technique is obtained when combined with the GLS and TS algorithm at a cost of 1769.8 and 7.01% improvement compared to the Two-Stage Algorithm. The lowest cost 200 initial customers and 150 dynamic customers when using the K-Means clustering with the PCA and GLS algorithm at a cost of 1906.9 and 8.25% improvement. The lowest cost 200 initial customers and 170 dynamic customers when using the K-Means clustering technique is with the PCA and GLS algorithm at a cost of 2092 and 3.75% improvement.

For large size problems, the lowest cost 300 initial customers and 220 dynamic customers when using the K-Means clustering technique is obtained when combined with the GLS and TS algorithm at a cost of 2275.9 and 8.28% improvement compared to the Two-Stage Algorithm. The lowest cost 300 initial customers and 250 dynamic customers when using the K-Means clustering with the PCA and GLS algorithm at a cost of 2410.4 and 4.37% improvement. The lowest cost 300 initial customers and 270 dynamic customers when using the K-Means clustering technique is with the GCA and GLS algorithm at a cost of 2539.3 and 5.27% improvement.

The disadvantage is that the construction algorithms may deliver an infeasible solution. In these cases, it is impossible to find an initial answer. Other construction algorithms that were tested delivered infeasible results in some cases. The third disadvantage of current methodology is updating the solution occurs after executing algorithms, which cause delays for starting and updating route for the vehicles.

As mentioned earlier, these three-stage algorithms have better outcomes than two-stage construction and improvement algorithms without the clustering. It means that the proposed methodology is an appropriate tool for LSDVRP. It may happen, some of the three-stage algorithms do not have any improvement over the 2-stage without clustering approach. One possible explanation for this discrepancy is that sometimes construction and improvement algorithms are enough to identify the best answer. Therefore, using clustering algorithms along with the construction and improvement heuristics typically does not make the solution worse.

LSDVRP is an NP-Hard problem. To achieve an exact solution for the LSDVRP is extremely hard and even impossible. Furthermore, heuristics with a limited run time may not find the global optimal solution. One of the advantages of the proposed three-stage heuristic is that it can solve this problem. On the other hand, the drawback of current method is that the obtained outcome may not be the global optimum answer.

In summary, this method is an effective way to solve large real-world LSDVRPs. Based on the data distribution, the best clustering algorithms are selected, and two-stage algorithms are run in each cluster to find the results. As a result, the outcomes indicate that the suggested heuristic approach provides a good result for the LSDVRPs.

## Conclusion and Future Work

The main objective of this research is to solve large size dynamic vehicle routing problems. In this paper, a hierarchical approach consisting of the cluster-first, construction-route second followed by an improvement route method is proposed. The first stage is used to cluster customers into ‘K’ groups based on the number of available vehicles (K), using three different algorithms (K-means, BIRCH, and GMM) separately. The second stage is the route construct for each cluster. The third stage is used to further improve the route using improvement algorithms. In addition, this methodology modifies the route whenever a new demand point enters the system.

Two main contributions are performed in this work: First, is the method of re-clustering in a LSDVRP when a new demand enters the system. Additionally, in this work, for the first time, a combination of the clustering algorithms and the construction and improvement algorithms is introduced. The proposed approach was tested in nine different case studies, and the results and improvement percentage were compared with the two-stage algorithm. The three-stage algorithms have better results than known algorithms as shown by comparison with two-stage algorithms that have been proposed earlier. Experimental analysis shows the ability of the algorithm to obtain the answer. Although the research did not conclusively identify the best sequence of algorithms, the three-stage procedure has improved the solution compared to the other two-stage algorithms.

The application of this technique is for solving LSDVRPs to find the best tours when the number of demand points is large. Also, the techniques can be used to solve different variants of the VRP and CVRP. Due to these favorable results, the proposed approach can be applied by companies to solve their VRPs.

Future research includes developing this methodology for uncertainty conditions, such as demand uncertainty. The assumption that the delivery demands of customers should be met in a soft and hard time window could be also explored. This methodology can apply to N-depots vehicle routing problems as well.

It is necessary to emphasize that the clustering of customers may lead to a loss of better solutions or even the best solutions. This is an acknowledged limitation of the study which was noticed during the execution of the algorithm. This approach may fail if the answer to the construction algorithm is infeasible. Therefore, it would be impossible to find an answer. The second limitation is that all the clustering algorithms might not be a good fit for the data set. Different data sets need different clustering algorithms that need to be verified. The major limitation of this study is that no good real dataset from a company is available to test this model. The heuristic could be tested with real data in the future. It is recommended that further studies using a large real data set be carried out.

# Appendix

**Fig. 5.** The best results of the proposed methodology by K-means, PCA and GLS

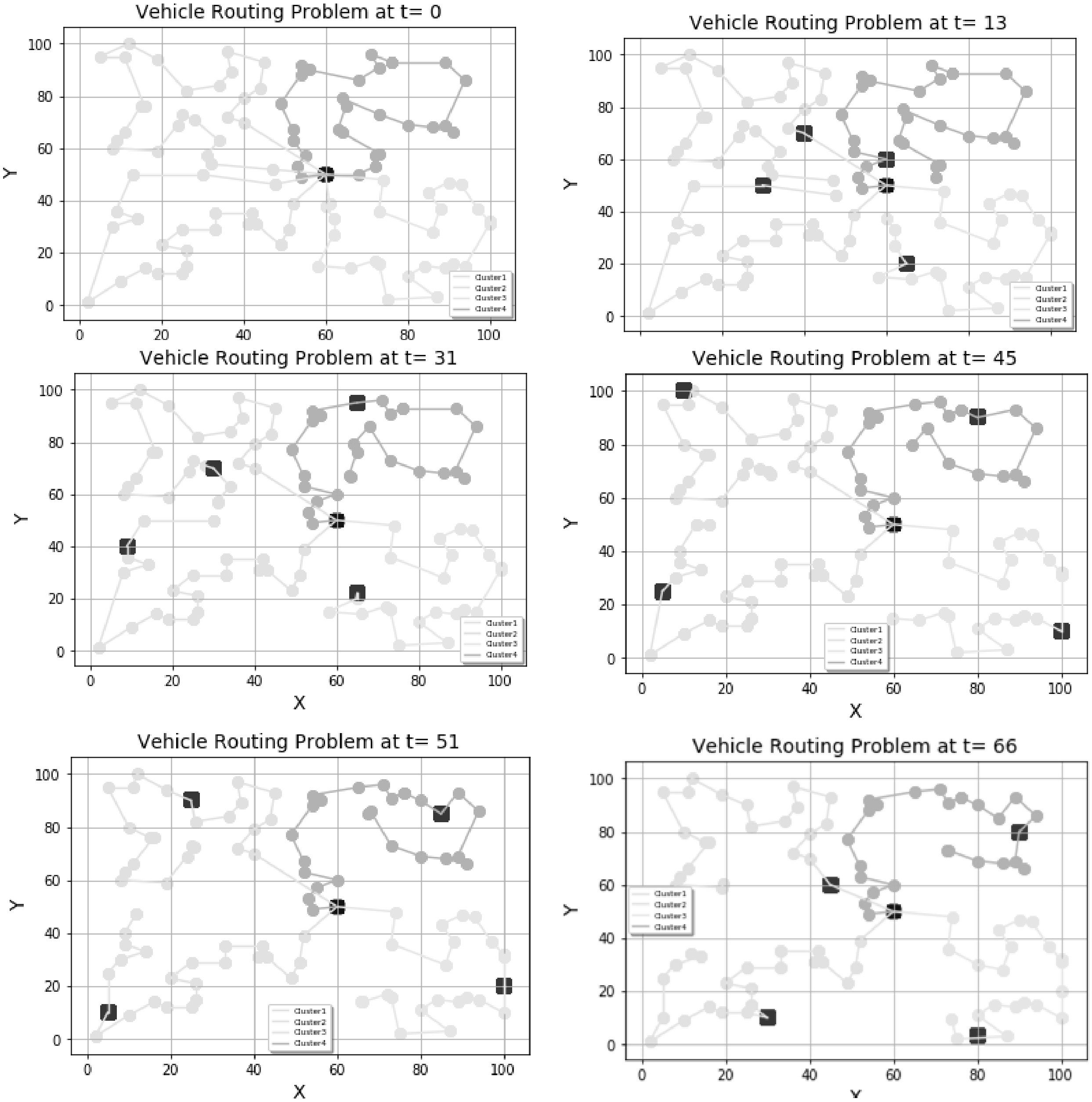